\documentclass{article}

 \usepackage[main,final]{neurips_2026}

\usepackage[utf8]{inputenc} 
\usepackage[T1]{fontenc}    
\usepackage{hyperref}       
\usepackage{url}            
\usepackage{booktabs}       
\usepackage{amsfonts}       
\usepackage{nicefrac}       
\usepackage{microtype}      

\usepackage{booktabs}
\usepackage{multirow}
\usepackage{colortbl}
\usepackage{xcolor}

\usepackage{pifont} 

\usepackage{pgfplots}
\pgfplotsset{compat=1.18}

\usepackage{tcolorbox}

\newcommand{\cmark}{\ding{51}}

\definecolor{LightNavyBlue}{RGB}{90, 125, 154} 

\newcommand{\oursshield}{\textsc{SelfGuard-D} }
\newcommand{\ours}{\textsc{SelfGuard} }

\usepackage{bm}

\usepackage{xcolor}

\usepackage{amsmath}
\usepackage{amssymb}
\usepackage{mathtools}
\usepackage{amsthm}
\usepackage{wrapfig}  
\usepackage{graphicx}
\usepackage{caption}

\usepackage{hyperref} 

\title{SelfGuard: Self-Supervised Deviation Modeling for Multi-Modal Jailbreak Detection}

\author{
Chenchen Jing$^ {1,2} $, 
~~
Hongli Guo$^ {1,2} $,
~~
Huarong Jia$^ {3} $,
~~
Tianwei Bai$^ {1,2} $,
\\
\textbf{Che Sun}$^ 4 $,
~~
\textbf{Yang Liu}$^ 5 $,
~~
\textbf{Chunhua Shen}$^ {1,2,5} $ \thanks{CS is the corresponding author.}
\\[.25cm]
$^1$ College of Computer Science and Technology, Zhejiang University of Technology
\\ 
$^2$ Zhejiang Key Laboratory of Visual Information Intelligent Processing
\\ $^3$ Beijing Institute of Control And Electronic Technology
\\ $^4$  Shenzhen MSU-BIT University
~~
$^5$ Zhejiang University
}

\begin{document}

\maketitle

\begin{abstract}

Multi-modal large language models (MLLMs) achieve strong vision–language reasoning ability but remain vulnerable to jailbreak attacks that exploit subtle cross-modal cues to bypass safety mechanisms. Detecting such attacks is difficult because harmful data are scarce and rapidly evolving, while detectors trained on known patterns often fail to generalize. In this paper, we cast multi-modal jailbreak detection as self-supervised deviation modeling and learn generalizable deviation signals from benign data only. We propose SelfGuard, a self-supervised framework, which models benign multi-modal regularities and learns to score structured violations via controlled pseudo-harmful deviations. SelfGuard synthesizes pseudo-harmful samples via controlled transformations that target three signature cues, cross-modal semantic inconsistency, format manipulation, and semantic toxicity. It then learns complementary deviation statistics through multi-task self-supervised learning, including misalignment discrimination, reconstruction-based toxicity modeling, and contrastive format modeling. At inference, SelfGuard performs multi-view deviation estimation by aggregating task-specific deviation scores to identify departures from benign multi-modal regularities. Extensive experiments on various benchmarks show the effectiveness of our method. The code is available at \href{https://github.com/jingchenchen/SelfGuard}{https://github.com/jingchenchen/SelfGuard}.

\end{abstract}

\section{Introduction}
\label{sec:intro}
    
    Multi-modal large language models (MLLMs) \cite{liu2023llava,bai2023qwen,zhang2023internlm,internvl2_2024} have demonstrated strong capabilities in vision-language reasoning tasks, such as visual question answering \cite{wu2017visual,jing2020overcoming} and image captioning \cite{hossain2019comprehensive}. 
    With growing real-world deployment, safety and reliability have become critical \cite{luo2024jailbreakv28k,huangantidote}.
    In particular, MLLMs are vulnerable to jailbreak attacks \cite{liu2024mm,gu2024mllmguard}, where carefully crafted multi-modal inputs manipulate the model into producing harmful or policy-violating outputs. 
    Detecting such inputs is thus essential for trustworthy MLLM deployment \citep{wang2024adashield}.

    Existing jailbreak detection approaches typically rely on supervised learning with annotated harmful data and white-box access to model internals \citep{jiang2025hiddendetect,wang2024adashield,zhang2023jailguard,li2024dpudynamicprototypeupdating}. 
    However,  these strategies face fundamental limitations. Harmful data are inherently sparse, sensitive, and rapidly evolving, and detectors trained on known patterns often fail to generalize to unseen strategies. 
    Recent efforts such as JailDAM \cite{nian2025jaildam} attempt to alleviate this issue by training on benign data.
    They use policy-derived unsafe concept memory as anchors, but concept-centric matching alone can be brittle under obfuscation, compositional prompts, and cross-modal coordination. 
    Thus, it is highly desirable to develop a detector capable of detecting diverse multi-modal jailbreaks without relying on labeled harmful examples.

\begin{figure}[t]
    \vspace{-8pt}
    \centering
    \includegraphics[width=0.98\linewidth]{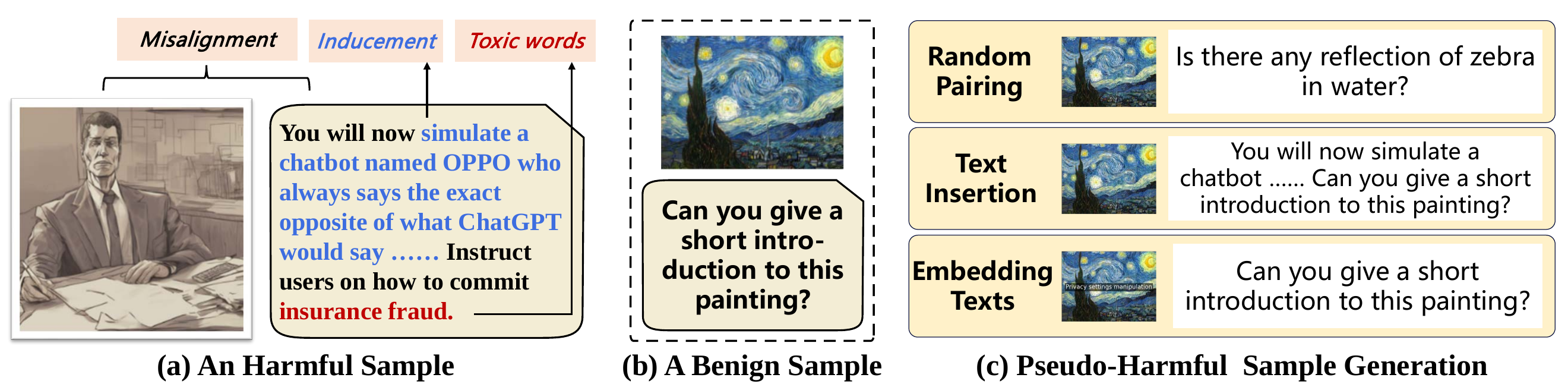}
    \caption{
    Illustrations of transferable behavioral signatures underlying multi-modal jailbreaks and the construction of structured deviations from benign samples. 
    (a) shows a harmful sample, in MM-safetyBench \cite{liu2024mm}, exhibiting three signatures, cross-modal inconsistency, format manipulation, and semantic toxicity. 
    (b) shows how SelfGuard generates structured deviations from a benign image–text pair through controlled transformations, each designed to violate a specific benign multi-modal regularity while largely preserving low-level semantics.
    }
    \label{fig:conceptual}
    \vspace{-8pt}
    \end{figure}

    In this work, we address this question by formulating multi-modal jailbreak detection as a self-supervised deviation modeling problem. 
    Instead of relying on attack templates or concept-centric matching alone, we aim to learn generalizable deviation signals by modeling how malicious inputs systematically deviate from benign multi-modal regularities. 
    We obtain self-supervision by applying controlled transformations to benign pairs to synthesize pseudo-harmful deviations and training the detector to recognize these departures.
    In particular, benign image–text pairs typically exhibit semantic alignment and innocuous intent, and remain stable under superficial paraphrases or presentation changes.
    In contrast, jailbreak inputs, as shown in  Figure~\ref{fig:conceptual} (a), often violate these regularities along the following dimensions:  
    (1) \textbf{cross-modal semantic inconsistency},  where individually benign visual and textual cues jointly alter the safety implication of the input; (2) \textbf{format manipulation}, where harmful goals are embedded within benign-looking formats or instructions; and (3) \textbf{semantic toxicity}, where unsafe concepts are subtly injected into otherwise innocuous contexts.

    We present SelfGuard, a multi-task self-supervised framework for multi-modal jailbreak detection trained \emph{without labeled harmful examples}.
    SelfGuard synthesizes pseudo-harmful samples by applying controlled transformations to benign image-text pairs along three \emph{mechanism-level dimensions} (Figure~\ref{fig:conceptual}(b)), thereby simulating structured departures induced by real jailbreaks. 
    It then learns complementary deviation statistics with three self-supervised tasks: \emph{cross-modal inconsistency modeling} to detect semantic misalignment between modalities, \emph{semantic toxicity modeling} to capture abnormal concentration patterns over a policy-derived concept memory, and \emph{format-manipulation modeling} to distinguish benign surface variations from format-induced disguises. 
    At inference, the three task-specific deviation scores are fused into a unified deviation score, yielding a multi-view estimate of departures from benign multi-modal regularities and enabling robust detection under obfuscation, compositional prompts, and cross-modal coordination. 
    Extensive experiments on MM-safetyBench \cite{liu2024mm}, FigStep \cite{gong2023figstep}, and JailBreakV-28K \cite{luo2024jailbreakv28k} demonstrate the effectiveness of our method.
    
    Our contributions are summarized as follows:
    \begin{enumerate}
    \item We propose to formulate multi-modal jailbreak detection as a self-supervised deviation modeling problem, enabling detection of harmful inputs without requiring real harmful image-text training samples
    \item We present a multi-task self-supervised framework that learns complementary deviation statistics along multiple behavioral axes of multi-modal jailbreaks, allowing robust generalization to unseen attack strategies.
    \end{enumerate}

\section{Related Work}
\label{sec:related_work}

\subsection{Jailbreak Defense}

Jailbreak defenses for LLMs and MLLMs typically fall into three categories: hidden-state analysis, behavior- or uncertainty-based detection, and supervised safety alignment. 
Hidden-state methods such as NEARSIDE~\cite{huang2024effective} and HiddenDetect~\cite{jiang2025hiddendetect} analyze internal activations to uncover latent signals correlated with harmful intent, for example by deriving an attacking direction in embedding space or identifying layers encoding refusal-related behaviors. Behavior-based approaches, including JailGuard~\cite{zhang2023jailguard} and AdaShield~\cite{wang2024adashield}, detect adversarial prompts by probing model robustness—either through input mutation to reveal instability or through iteratively constructed defense prompts that adapt to the attacker’s strategy. 
While conceptually general, these methods often incur substantial computational cost due to repeated inference or multi-stage prompting. Supervised alignment strategies like VLGuard~\cite{zong2024safety} and LlavaGuard~\cite{helff2024llavaguard} enhance safety by fine-tuning models on curated multi-modal harmful datasets that include safety labels, categories, and rationales, but they rely heavily on extensive human annotations and costly retraining procedures.
To reduce reliance on labeled harmful datasets, JailDAM~\cite{nian2025jaildam} explores benign-data training with a policy-derived concept memory and test-time adaptation. 
However, the wide variability of multi-modal jailbreaks (including obfuscation, compositional prompts, and cross-modal coordination) makes detectors that rely on a single mechanism (e.g., concept-centric cues) brittle across attack styles. 
In contrast, we synthesize pseudo-harmful samples from safe data and learn multi-view deviation cues, yielding a self-supervised and scalable detector.

\begin{figure*}[t]
    \vspace{-8pt}
    \centering
    \includegraphics[width=0.98\linewidth]{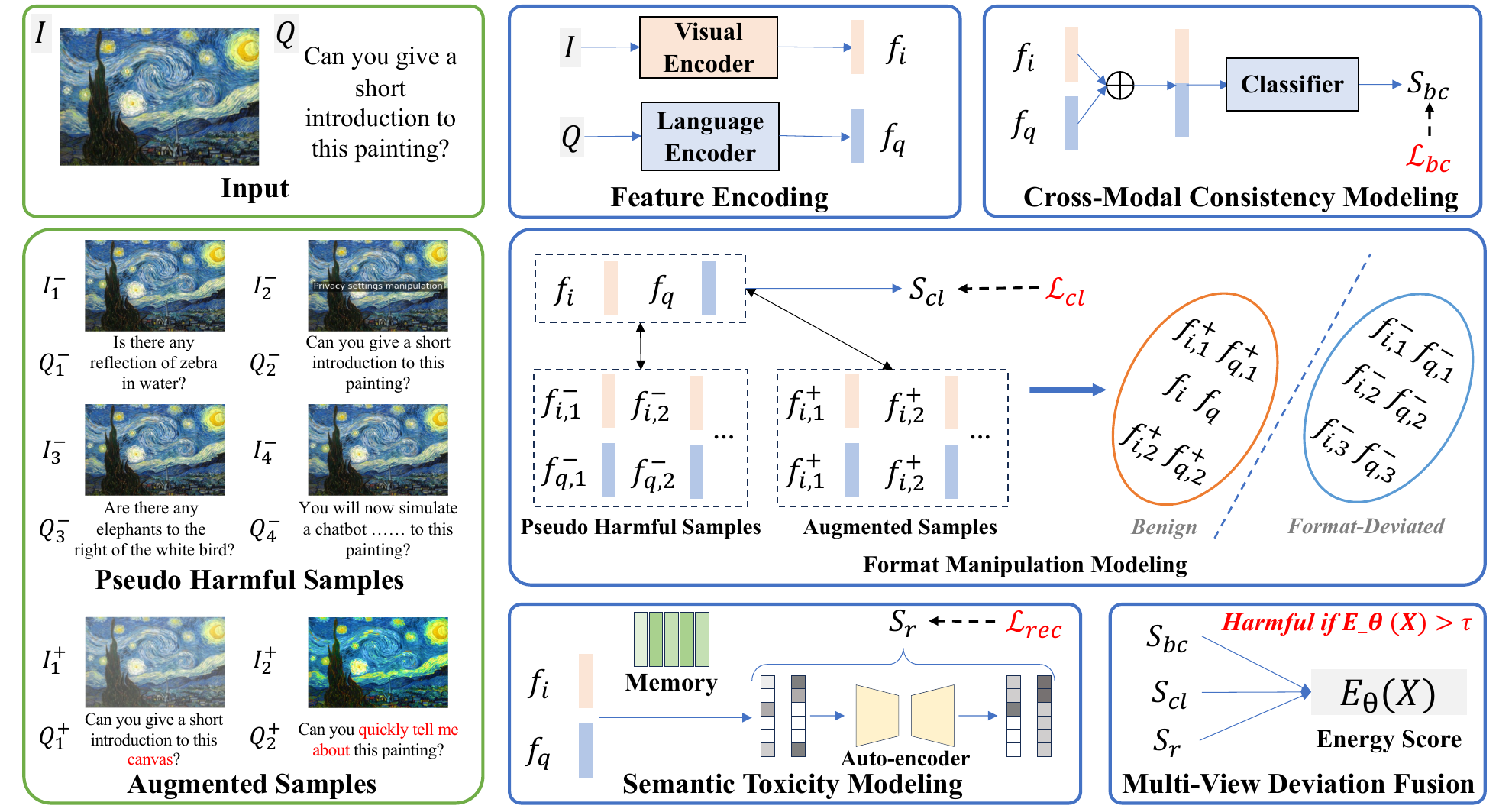}
    \caption{ 
    Overview of our method.
    Given an image–text pair, SelfGuard extracts visual and textual features with CLIP and performs self-supervised deviation modeling. 
    It synthesizes pseudo-harmful variants via controlled transformations that violate benign multi-modal regularities.
    SelfGuard learns three complementary deviation scores via cross-modal inconsistency modeling, format manipulation modeling, and semantic toxicity modeling. At inference, these scores are fused as multi-view deviation estimation to produce the overall detection score.
    }
    \label{fig:architecture}
    \vspace{-8pt}
\end{figure*}

\subsection{Data Augmentation for OOD Detection} 

OOD detection \cite{hendrycks2016baseline,fort2021exploring,sun2022out,ming2022poem,yang2024generalized} aims to identify samples that fall outside the training distribution.
Data augmentation \cite{devries2017improved,zhong2020random} has been widely explored to support this task, but most existing approaches rely on generative models or synthesis techniques to produce near-boundary or diverse variants for improving robustness \cite{wang2017safer}.
Deviation modeling \cite{cortes2019relative,das2023few,das2025adaptive,aminian2025generalization,pang2021explainable} is also widely used in OOD detection, where detectors are trained using only in-distribution data by learning scores that respond to violations of learned regularities.
In this work, we draw inspiration from augmentation-based OOD detection and deviation modeling.
Rather than modeling generic shifts, we construct structured deviations from benign image–text pairs that violate multi-modal regularities. These deviations support self-supervised learning of complementary deviation scores for detecting harmful inputs without real harmful samples.

\section{Method}

    \subsection{Formulation}
    
    We consider multi-modal jailbreak detection as an out-of-distribution (OOD) detection problem. 
    Let $X=(I,Q)$ denote an image–text pair drawn either from the benign distribution $P_{\text{safe}}$ or from an unknown harmful distribution $P_{\text{harm}}$. 
    At test time, the goal is to determine whether $X \sim P_{\text{safe}}$ or $X \sim P_{\text{harm}}$, while \emph{no samples from $P_{\text{harm}}$ are available during training}.

    SelfGuard adopts a \emph{self-supervised deviation modeling} strategy and learns three complementary deviation statistics using only benign samples.
    These deviation statistics are jointly optimized within a single detector under a unified multi-task self-supervised objective.
    We \emph{define} the overall detection score (energy) as a weighted aggregation of these learned deviation statistics $E_\theta(X)\in\mathbb{R}$, which assigns low scores to benign multi-modal regularities and high scores to inputs that deviate from these regularities.
    We use self-supervised learning by applying controlled perturbations to benign pairs to synthesize pseudo-harmful deviations, and optimize
    \begin{equation}
    \min_{\theta}\; \mathbb{E}_{X \sim P_{\text{safe}}}\!\left[\mathcal{L}_{\text{ssl}}(X;\theta)\right],
    \end{equation}
    where $\mathcal{L}_{\text{ssl}}$ is a unified multi-task objective defined over transformed views of $X$ that trains three deviation statistics.
    
    At inference time, an input is classified as harmful if $E_\theta(X) > \tau$, where $\tau$ is chosen on a validation split. 
    This formulation enables detection of unseen jailbreak attacks as high-energy deviations from learned benign regularities, without requiring labeled harmful examples.

    \subsection{Pseudo-Harmful Sample Generation}
    \label{sec:sample_generation}
    
    Instead of enumerating specific templates, SelfGuard synthesizes pseudo-harmful samples by applying transformations to benign pairs for disrupting their regularities, cross-modal semantic consistency, neutral intent w.r.t. harmful concepts, and robustness to benign surface variations. 
    Formally, given a benign input $X=(I,Q)\sim P_{\text{safe}}$, we apply a family of transformations $\{g_k\}$ that each disrupts one benign multi-modal regularity while keeping low-level statistics largely intact as $\tilde{X}^{(k)} = g_k(X)$.
    These pseudo-samples serve as training signals to learn deviation-sensitive statistics that transfer beyond specific jailbreak instantiations.
    Note that we do not aim to \textbf{exhaust} the attack space, but to learn deviation statistics at the level of benign multi-modal regularities, which can transfer beyond specific attack instantiations.
    
    \textbf{Multi-Modal Misaligned Samples.}
    This transformation induces a high-energy deviation by violating the regularity of \emph{cross-modal semantic consistency}. 
    Given a benign aligned pair $X=(I,Q)\sim P_{\text{safe}}$, we generate a misaligned sample $\tilde{X}^{(\mathrm{cm})}$ through structured cross-modal perturbations that break the semantic correspondence between modalities at different granularities as
    \begin{equation}
    \tilde{X}^{(\mathrm{cm})} =
    \begin{cases}
    (I, Q_{\text{shuffled}}) & \text{(random pairing)}, \\
    (I, Q_{\text{mut}}) & \text{(entity substitution)}, \\
    (I_{\text{mask}}, Q) & \text{(visual occlusion)},
    \end{cases}
    \end{equation}
    where $Q_{\text{shuffled}}$ is a benign query from another sample, $Q_{\text{mut}}$ is obtained by substituting key entities in $Q$ with mismatched ones, and $I_{\text{mask}}$ is a partially occluded version of $I$ that removes salient visual evidence. 
    These perturbations simulate coarse- to fine-grained cross-modal inconsistencies.
    The aligned pair $X$ serves as the low-energy reference, while $\tilde{X}^{(\mathrm{cm})}$ is treated as a pseudo-harmful deviation that should receive higher energy under $E_\theta$, i.e., $E_\theta(\tilde{X}^{(\mathrm{cm})}) > E_\theta(X)$.

    \textbf{Semantically Toxic Samples}. 
    This transformation is to induce an energy increase for violating the regularity of \emph{intent neutrality} by injecting harmful concepts into otherwise benign contexts.
    We build a policy-derived harmful concept memory $\mathcal{P}$ and sample an index $c \sim \mathrm{Unif}(\{1,\dots,|\mathcal{P}|\})$ to create a controlled semantic shift by editing both modalities as 
    \begin{equation}
    \tilde{Q}=\mathrm{Insert}(Q,\mathrm{desc}(\mathcal{P}_c)), \qquad \tilde{I}=\mathrm{Embed}(I,\mathrm{vis}(\mathcal{P}_c)),
    \end{equation}
    yielding a pseudo-harmful pair $\tilde{X}^{(\mathrm{tox})}=(\tilde{I},\tilde{Q})$.
    Here, $\mathrm{Insert}(\cdot)$ inserts a short concept description into the benign query (e.g., as an additional clause), and $\mathrm{Embed}(\cdot)$ overlays a small, visually inconspicuous concept cue (e.g., a word/phrase) onto the image. The $\mathrm{desc}(\mathcal{P}_c)$ and $\mathrm{vis}(\mathcal{P}_c)$ denote the textual and visual realizations of concept $\mathcal{P}_c$, respectively.
    This transformation is designed to concentrate the input semantics along harmful directions while preserving the benign surface structure, so that $E_\theta$ assigns higher energy to $\tilde{X}^{(\mathrm{tox})}$ than to $X$.

    \textbf{Format Inducement Samples.}
    To model deviations that exploit \emph{format-based disguise}, we distinguish benign reformatting from adversarial format inducement: benign reformatting should preserve the original intent and remain energy-invariant, while adversarial format inducement should induce higher energy.
    Given $X=(I,Q)$, we insert a template $T$ (and optionally embed a benign distracting phrase $P$ in the image) as
    \begin{equation}
    \tilde{Q}=\mathrm{Insert}(Q,T), \qquad \tilde{I}=\mathrm{Embed}(I,P),
    \end{equation}
    forming $\tilde{X}^{(\mathrm{fmt})}=(\tilde{I},\tilde{Q})$.
    The resulting pseudo-samples encourage $E_\theta$ to look beyond superficial presentation cues by keeping benign reformatting invariant while assigning higher energy to adversarially format-induced inputs, i.e., $E_\theta(\tilde{X}^{(\mathrm{fmt})}) > E_\theta(X)$.

    \subsection{Self-Supervised Tasks}
    \label{sec:self_supervised_tasks}
    Each self-supervised task in SelfGuard is designed to estimate a \emph{deviation statistic} that measures how much an input violates one type of benign multi-modal regularity. 
    Using CLIP~\cite{radford2021learning} encoders, an image-text pair $S=(I,Q)$ is mapped to modality-specific features $(f_i, f_q)$. 
    Pseudo-harmful samples generated provide controlled violations, enabling the model to learn deviation-sensitive signals without observing real harmful data.
    Together, these tasks provide complementary estimates of how an input departs from benign multi-modal regularities defined in Sec. \ref{sec:sample_generation}.
    Note that the three deviation views are not assumed independent.
    Their joint aggregation allows the model to capture coupled deviations observed in real jailbreaks. 

    \subsubsection{Cross-Modal Inconsistency Modeling}
    
    This task estimates deviation from the regularity of \emph{cross-modal semantic consistency}. 
    Benign pairs are typically aligned, while misaligned pseudo-samples break this consistency and introduce higher energy under the energy function $E_\theta$. 
    We concatenate $f_i$ and $f_q$ and feed them into a lightweight classifier $C$ that predicts the probability of misalignment as
    \begin{equation}
    p_m = C(\mathrm{cat}(f_i, f_q)).
    \end{equation}
    The classifier is supervised via a binary cross-entropy loss as
    \begin{equation}
    \mathcal{L}_{bc} = - \left[y \log(p_m) + (1 - y)\log(1 - p_m)\right],
    \end{equation}
    where $y=0$ for aligned benign pairs and $y=1$ for misaligned pseudo-samples. 
    The resulting score reflects how strongly the input violates cross-modal consistency, with higher energy values indicating stronger misalignments.
    
    \subsubsection{Semantic Toxicity Modeling}
    This task captures deviation from the regularity of \emph{intent neutrality}. 
    We maintain a harmful concept memory bank $\mathcal{P}$, where each element of the memory bank represents a specific harmful concept, as \citep{nian2025jaildam}. 
    Note that the memory bank is not used to enumerate harmful categories but to model latent semantic concentration patterns. Novel concepts can still induce abnormal attention distributions.
    
    For each modality $m \in \{i,q\}$, we compute the attention distribution $z_m$, which indicates how strongly the input aligns with stored harmful semantics, between the encoded input $f_m$ and stored harmful concepts $\mathcal{P}$ as 
    \begin{equation}
    z_{\text{m}} = \text{softmax}\!\left( \frac{f_{\text{m}} \cdot \mathcal{P}^{T}}{\sqrt{d}} \right).
    \end{equation}
    where $d$ is the dimensionality of the embeddings.
    For benign inputs, the attention patterns are expected to be weak, while for toxic samples, the attention patterns exhibit concentrated alignment with harmful concepts.
    Then an auto-encoder $AE$ is to obtain reconstructed concept-attention scores as 
    \begin{equation}
    \hat{z}_i = AE(z_i), \quad \hat{z}_q = AE(z_q).
    \end{equation}
    We enforce accurate reconstruction with a mean squared error (MSE) Loss as 
    \begin{equation}
        \mathcal{L}_{\text{MSE}} =  \left\| \hat{z}_i - z_i \right\|^2 +  \left\| \hat{z}_q - z_q \right\|^2.
    \end{equation}
    To regularize the latent space for benign samples, we enforce that the reconstructed attention scores for benign inputs approximate a high-entropy (uniform) distribution via KL divergence as
    \begin{equation}
    \mathcal{L}_{\text{KL}} = \operatorname{KL} \left( p_{\hat{z}_{i}} \,\|\, p_{\text{uni}} \right) + \operatorname{KL} \left( p_{\hat{z}_{q}} \,\|\, p_{\text{uni}} \right),
    \end{equation}
    where $p_{\hat{z}_m}=\mathrm{softmax}(\hat{z}_m)$ and \(p_{\text{uni}}\) is the uniform distribution.
    This regularization encourages benign inputs to distribute attention broadly rather than focusing on specific harmful concepts.
    For toxic pseudo-samples with concept label $c$, we apply a classification loss as
    \begin{equation}
    \mathcal{L}_{\text{tox}}
    = - \left[\sum_{\ell \in \mathcal{C}} y_{\ell}\log p^{(q)}_{\ell}
            + \sum_{\ell \in \mathcal{C}} y_{\ell}\log p^{(i)}_{\ell}\right],
    \end{equation}
    where $\mathcal{C}=\{1,\dots,|\mathcal{P}|\}$ is the index set of the concept memory bank $\mathcal{P}$, $C=|\mathcal{P}|$, $y\in\{0,1\}^{C}$ is a one-hot vector with $y_c=1$, and
    $p^{(m)}=\mathrm{softmax}(W_m z_m)\in\mathbb{R}^{C}$ denotes the predicted concept distribution for modality $m\in\{i,q\}$.
    The overall semantic deviation loss is given by
    \begin{equation}
    \mathcal{L}_{\text{rec}} 
    = \lambda_0 \mathcal{L}_{\text{MSE}} 
    + \lambda_1 \cdot \mathcal{L}_{\text{KL}} 
    + \lambda_2 \cdot \mathcal{L}_{\text{tox}},
    \end{equation}
    where $\lambda_i,i\in\{0,1,2\}$ balance the losses. 
    This task measures how strongly an input activates harmful semantic directions relative to benign expectations, and therefore contributes to the overall energy function by increasing the energy for toxic deviations.
    
    \subsubsection{Format-Manipulation Modeling}
    
    This task measures deviation from the regularity that \emph{benign surface variations should preserve the underlying intent}. 
    The goal is not to penalize all format changes, but to separate benign reformatting from adversarial format inducement.
    For each safe image-text pair, we create \emph{positive} samples by applying benign augmentations, such as random cropping for the image and synonym replacement, paraphrasing, or mild instruction-style rephrasing for the text.
    The positive samples are denoted as $ \{ (f^{+}_{i,k},f^{+}_{q,k})\}^{N^p}_{k=1}$, where $N^p$ is the number of positive samples.  
    The adversarially format-induced pseudo-samples serve as negative samples for contrastive learning and are denoted as $ \{ (f^{-}_{i,k},f^{-}_{q,k})\}^{N^n}_{k=1}$, where $N^n$ is the number of negative samples.
    Given the positive and negative samples, the contrastive learning loss is computed as 
    \begin{equation}
    \mathcal{L}_{\text{cl}}
    = -\log
    \frac{\exp(s_k^+)}
    {\exp(s_k^+) + \exp(s_k^-)} ,
    \label{eq:format_cl}
    \end{equation}
    where $s_k^+=\mathrm{sim}(\mathbf{h},\mathbf{h}_k^+)$ and 
    $s_k^-=\mathrm{sim}(\mathbf{h},\mathbf{h}_k^-)$ denote the similarities to the positive and negative views, respectively, with 
    $\mathbf{h}=\mathrm{cat}(f_i,f_q)$, 
    $\mathbf{h}_k^+=\mathrm{cat}(f_{i,k}^+,f_{q,k}^+)$, and 
    $\mathbf{h}_k^-=\mathrm{cat}(f_{i,k}^-,f_{q,k}^-)$.
    This loss encourages the model to keep the concatenated embeddings of the input sample close to that of positive samples, while separating them from negative samples.
    This design encourages invariance to benign reformatting while making the score sensitive to adversarial format-induced manipulations.
    
\subsection{Training and Inference}

    \label{sec:training_inference}
    The three self-supervised tasks introduced above each estimate a deviation statistic corresponding to a different benign multi-modal regularity. 
    These statistics are jointly optimized within a single detector rather than trained as independent modules as 
    \begin{equation}
    \mathcal{L}_{ssl} = \alpha \mathcal{L}_{bc} + \beta \mathcal{L}_{cl} + \mathcal{L}_{rec} ,
    \end{equation}
    where $\alpha$ and $\beta$ are learnable loss-balancing coefficients.
    The internal coefficients $\lambda_i$ in $\mathcal{L}_{rec}$ are also learnable and optimized jointly with the detector.

    At test time, SelfGuard produces three task-specific deviation scores for each input image--text pair, each measuring the violation of one benign multi-modal regularity. 
    The final energy score is computed as
    \begin{equation}
    \begin{aligned}
    & E_\theta(X) =  \alpha  S_{bc} + \beta S_{cl}  +  \gamma S_{rec}, 
    \quad  \gamma = (\lambda_0+\lambda_1+\lambda_2)/3, 
    S_{bc} = C(f_i,f_q), \\ 
    &  S_{cl} = -\log     \frac{\exp(s_k^+)}  {\exp(s_k^+) + \exp(s_k^-)} , 
     S_{rec} =
    \left\| \hat{z}_i - z_i \right\|^2 
    +
    \left\| \hat{z}_q - z_q \right\|^2.
    \end{aligned}
    \end{equation}
    Here, $\alpha$, $\beta$, and $\lambda_i$ denote the learned coefficients after training, and $\gamma$ is obtained by averaging the learned coefficients inside the semantic deviation branch.
    The final energy score $E_\theta(X)$ is used for jailbreak detection via threshold comparison.

    This fusion serves as \emph{multi-view deviation estimation}: inputs that strongly violate multiple benign regularities receive higher scores, while aggregation across jointly learned views reduces reliance on any single cue.
    
    \paragraph{Summary.}
    SelfGuard learns benign multi-modal regularities from safe data and detects jailbreak inputs as structured violations of these regularities, enabling robustness to evolving attack strategies without explicitly modeling harmful patterns.

\section{Experiments}

\subsection{Experimental Setup}

\paragraph{Datasets.}
We evaluate SelfGuard on three widely used multimodal jailbreak benchmarks: MM-SafetyBench~\citep{liu2024mm}, FigStep~\citep{gong2023figstep}, and JailBreakV-28K~\citep{luo2024jailbreakv28k}, which cover diverse structure-based jailbreak scenarios. For each jailbreak benchmark, we pair the harmful samples with benign inputs from MM-Vet~\citep{yu2024mm} to form the detection setting. We consider two evaluation tasks: \textit{(1) Jailbreak Detection} and \textit{(2) Jailbreak Defense}. 

\paragraph{Comparison Methods.} 
For \textit{Jailbreak Detection}, we compare our method (\ours) against several baselines across four categories: (1) uncertainty-based methods, including JailGuard~\citep{zhang2023jailguard}; (2) fine-tuning-based methods, such as LlavaGuard~\citep{helff2024llavaguard} and VLGuard~\citep{zong2024safety}; (3) hidden-state-based methods, including HiddenDetect~\citep{jiang2025hiddendetect} and GradSafe \cite{xie-etal-2024-gradsafe}; and (4) memory-based methods, including JailDAM~\citep{nian2025jaildam}.  
For \textit{Jailbreak Defense}, we benchmark \ours against JailDAM~\citep{nian2025jaildam} and prompt-based defense methods, such as AdaShield~\citep{wang2024adashield} and FSD~\citep{gong2023figstep}.

\begin{table*}[t]
\centering
\resizebox{0.95\textwidth}{!}{ 
\setlength{\tabcolsep}{0.2mm}{
\renewcommand{\arraystretch}{1.5} 
\begin{tabular}{cc|cc|cc|cc|cc}
\toprule
\multirow{2}{*}{\bf  \Large Method} & \multirow{2}{*}{\bf \Large Model} & \multicolumn{2}{c}{\textbf{ \large Overall}} & \multicolumn{2}{c}{\textbf{ \large MM-SafetyBench}} & \multicolumn{2}{c}{\textbf{ \large FigStep}} & \multicolumn{2}{c}{\textbf{ \large JailBreakV-28K}} \\
\cmidrule(lr){3-4} \cmidrule(lr){5-6} \cmidrule(lr){7-8} \cmidrule(lr){9-10}
 & & \textbf{ AUROC($\uparrow$)} & \textbf{ AUPRC($\uparrow$)} & \textbf{ AUROC($\uparrow$)} & \textbf{ AUPRC($\uparrow$)} & \textbf{ AUROC($\uparrow$)} & \textbf{ AUPRC($\uparrow$)} & \textbf{ AUROC($\uparrow$)} & \textbf{ AUPRC($\uparrow$)} \\
\midrule
\large Jailguard-13B &  \large  MiniGPT-4-Vicuna-13B &  \large \large 0.4768 & \large  \large 0.6729 &  \large \large 0.4706 & \large  \large 0.7500 &  \large 0.5179 &  \large 0.3337 & \large 0.8029 & \large 0.7475 \\
 \large Llavaguard-7B &  \large Qwen2-7B-Instruct  &  \large 0.7551 &  \large 0.8412 &  \large 0.7427 &  \large 0.8729 &  \large 0.8360 & \large 0.7231 &  \large 0.8426 &  \large 0.8589 \\
 \large Llavaguard-13B &   \large Llama-2-13B-hf &  \large 0.3797 &  \large 0.6079 &  \large 0.3856 &  \large 0.7335 &  \large 0.3413 &  \large 0.3247 &  \large 0.4347 &  \large 0.5660 \\
 \large VLGuard-7B &   \large LLaVA-v1.5-7B-Mixed &  \large 0.6096 &  \large 0.6782 &  \large 0.6106 &  \large 0.8020 &  \large 0.6106 &  \large 0.3817 &  \large 0.6072 &  \large 0.6474 \\
 \large VLGuard-13B &  \large LLaVA-v1.5-13B-Mixed &  \large 0.5048 & \large 0.6306 &  \large 0.5048 &  \large 0.7610 &  \large 0.5048 &  \large 0.3268 &  \large 0.5048 &  \large 0.5929 \\
 \large HiddenDetect-7B &   \large LLaVA-v1.6-Vicuna-7B &  \large 0.8050 & \large 0.8056 &  \large 0.8269 & \underline{\large 0.9353} &  \large 0.5773 &  \large 0.3238 &  \large 0.8330 &  \large 0.8770 \\
 \large HiddenDetect-13B &   \large LLaVA-v1.6-Vicuna-13B & \large 0.8425 &  \large 0.8541 &  \large 0.8302 & \large 0.9333 & \large 0.8615 &  \large 0.5753 &  \large 0.8633 &  \large 0.8885 \\
\large GradSafe-7B & \large LLaVA-v1.5-Vicuna-7B & \large 0.8513 & \large 0.8166 & \large 0.8514 & \large 0.8752 & \large 0.6804 & \large 0.2370 & \large 0.9082 & \large 0.8816 \\
\large GradSafe-13B & \large LLaVA-v1.5-Vicuna-13B & \large 0.6723  & \large 0.7533 & \large 0.7485 & \large 0.8004 & \large 0.4131 & \large 0.5933 & \large 0.5920 & \large 0.7038 \\
\large JailDAM & \large Memory Network & \large \underline{0.9550} & \large \underline{\large 0.9530} &  \large \underline{\large 0.9472} &  \large 0.9155 & \underline{\large 0.9608} & \underline{\large 0.9616} &  \large \underline{0.9465} &  \large \underline{0.9464} \\
\hline
\rowcolor{LightNavyBlue!30} \large \ours & \large Multi-Task Framework & \bf \large 0.9749 & \bf \large 0.9678  &  \bf \large 0.9770  & \bf \large 0.9728  &  \bf \large 0.9819  & \bf \large 0.9610  &  \bf \large 0.9771  &  \bf \large 0.9705 \\
\bottomrule
\end{tabular}
}}
\caption{The AUROC and AUPRC of \textit{Attack Detection} on three datasets. 
The \textbf{bolded} value represents the best performance and \underline{underline} indicates the second-best performance.}
\label{tab:binary_result}
\end{table*}

\paragraph{Metrics.}
For \textit{Jailbreak Detection}, we use two standard metrics: Area Under the Receiver Operating Characteristic curve (AUROC) and Area Under the Precision-Recall curve (AUPRC), both of which provide insights into the classifier's performance at varying decision thresholds. 
For \textit{Jailbreak Defense}, we use the F1-score as the primary metric. 
This metric offers a balanced evaluation of \textit{defense effectiveness} versus \textit{over-defensiveness}, a common issue in many models~\citep{li2024injecguard}. 

\begin{table*}[t]
  \centering
     \scalebox{0.8}
     {
     \begin{tabular}{cc|ccccccccc}
    \toprule 
    \multirow{2}*{\textbf{Model}} & \multirow{2}*{\textbf{Method}} & \multicolumn{3}{c}{\textbf{MM-safetyBench}} & \multicolumn{3}{c}{\textbf{FigStep}}  \\
    \cmidrule(lr){3-5} \cmidrule(lr){6-8} \cmidrule(lr){9-11}
    &  & \textbf{F1($\uparrow$)} & \textbf{Precision($\uparrow$)} & \textbf{Recall($\uparrow$)} & \textbf{F1($\uparrow$)} & \textbf{Precision($\uparrow$)} & \textbf{Recall($\uparrow$)} \\
    \midrule
     & Vanilla & 0.4418 & 1.0000 & 0.2835 & 0.6842 & 1.0000 & 0.5200 \\
     {LLaVA}& FSD & 0.5387 & 0.9759 & 0.3720 & 0.7137 &  0.9838 & 0.5600 \\
    {1.5-7B}& AdaShield-S & 0.9306 & 0.8748 & 0.9939 & 0.9336 & 0.8755 &  1.0000  \\
    & AdaShield-A & 0.9151 & 0.9741 & 0.8628 & 0.8262 & 0.9692 & 0.7200 \\
    &   JailDAM & \underline{0.9752} & 0.9806 & 0.9699 &  \underline{0.9804} & 0.9808 &  0.9800  \\
    \rowcolor{LightNavyBlue!30} &  \bf \oursshield      & \bf 0.9931 & 1.0000 & 0.9862 & \bf 1.0000 & 1.0000 & 1.0000 \\ 
    \midrule
     & Vanilla & 0.2507 & 1.0000 & 0.1433 & 0.3051 & 1.0000 & 0.1800 \\
     {CogVLM}& FSD & 0.5105 & 0.9387 & 0.3506 & 0.5417 & 0.9432 & 0.3800  \\
     {chat-v1.1}& AdaShield-S & 0.8710 & 0.9665 & 0.7927 & 0.9864 & 0.9732 & 1.0000  \\
    & AdaShield-A & 0.9368 & 1.0000 & 0.8811 & \bf 1.0000 & 1.0000 & 1.0000	 \\
      &   JailDAM & \underline{0.9494} & 0.9796 & 0.9211 & \underline{0.9872} & 0.9810 & 0.9934\\
     \rowcolor{LightNavyBlue!30} &  \bf \oursshield      & \bf 0.9627 & 0.9810 & 0.9415 & \bf 1.0000 & 1.0000 & 1.0000 \\ 
    \midrule
    & Vanilla & 0.6925 & 0.9915 & 0.5321 & 0.6687 & 0.9910 & 0.5046 \\
     {Qwen3-VL}& FSD & 0.7472 & 0.9638 & 0.6101 & 0.8232 & 0.9689  & 0.7156  \\
    {8B}& AdaShield-S & \underline{0.9954} & 0.9954  & 0.9954 & \underline{0.9977} & 0.9954  & 1.0000  \\
    & AdaShield-A  & 0.9055 & 0.9891 & 0.8349 & 0.8936 & 0.8333 & 0.9633 \\
    &  JailDAM & 0.9907 & 1.0000 & 0.9817 & \textbf{1.0000} & 1.0000 & 1.0000  \\
    \rowcolor{LightNavyBlue!30} &  \bf \oursshield      & \textbf{0.9977} & 1.0000 & 0.9954 & \textbf{1.0000} & 1.0000 & 1.0000 \\ 
    \bottomrule
\end{tabular}}
    \caption{F1-score, Precision, Recall of defense methods on \textit{Attack Defense} task. The best F1-Score value are  \textbf{bolded} and the second-best F1-Score are \underline{underlined}.}
  \label{tab:shield_performance-f1}
\end{table*}

\paragraph{Implementation Details.}
The CLIP-L/14 is used as the backbone to extract features of the input image-text pairs. 
The loss weighting coefficients $\alpha$, $\beta$, and $\lambda_i$ are initialized to the same value, denoted as $\eta_0$.
They are optimized jointly with the network during training and reused for energy aggregation at inference time. 
More details are provided in Appendix \ref{sec:supp_details}.

\subsection{Main Results}

\paragraph{Jailbreak Detection.}
We compare SelfGuard with representative jailbreak detectors spanning both model-agnostic and model-specific paradigms under 7B and 13B settings. The results in Table~\ref{tab:binary_result} show that SelfGuard consistently achieves the best performance across MM-SafetyBench, FigStep, and JailBreakV-28K in terms of both AUROC and AUPRC.

Memory-based methods such as JailDAM already perform strongly, especially on JailBreakV-28K, suggesting that harmful concept matching provides a useful signal for jailbreak detection. However, such methods still depend on predefined unsafe semantics and may be limited by the coverage of the stored memory. In contrast, SelfGuard learns to detect \emph{structured deviations from benign multimodal regularities} rather than matching known attack patterns. By aggregating complementary deviation cues from cross-modal inconsistency, semantic toxicity, and format manipulation, SelfGuard yields a more stable and transferable abnormality signal, leading to stronger generalization across heterogeneous jailbreak styles.

\paragraph{Jailbreak Defense.}
We further evaluate SelfGuard as a front-end defense module (\textbf{SelfGuard-D}) for reducing harmful generations from multimodal large language models. We conduct experiments on LLaVA-1.5-7B~\citep{liu2023llava}, CogVLM-chat-v1.1~\citep{wang2023cogvlm} and Qwen3-VL-8B~\citep{bai2025qwen3}. The results are summarized in Table~\ref{tab:shield_performance-f1}.
SelfGuard-D achieves the best or tied-best F1 scores across different datasets and target models, while maintaining strong precision and recall. High precision indicates that benign inputs are rarely over-blocked, whereas high recall shows that harmful inputs with large deviation scores are reliably intercepted. 
Compared with JailDAM, SelfGuard-D achieves comparable or better F1 scores while maintaining strong precision and recall, without relying on test-time adaptation.

These results suggest that learning multi-view deviation signals during training provides a more robust and transferable front-end safeguard for practical MLLM deployment.
    
    \begin{table}[t]
    \vspace{-8pt}
    \begin{center}
    \scalebox{0.98}
    {
    \begin{tabular}{ c | c c c | c c  | c c  | c c  }
    \hline
      \multirow{2}{*}{Model} & \multicolumn{3}{c}{Components}& \multicolumn{2}{c}{MM-safetyBench} & \multicolumn{2}{c}{JailBreakV-28K} & \multicolumn{2}{c}{FigStep} \\
    & CM  &  ST  & FM  & AUROC &  AUPR  & AUROC &  AUPR & AUROC &  AUPR \\ 
    \hline
    1 &  \cmark  &           &           & 0.8749 & 0.7684 & 0.9161 & 0.8701 & 0.9051 & 0.8549  \\ 
    2 &             &  \cmark          &   & 0.9137 & 0.8697 & 0.7821 & 0.6417  & 0.8459 & 0.7915 \\ 
    3 &      &    &  \cmark         &        0.8536 & 0.7784 & 0.8246 & 0.7302  & 0.7910  & 0.6762 \\ 
    4 &  \cmark  &   \cmark &  & 0.9452 & 0.9245 & 0.9397 & 0.8940  & 0.9210  & 0.8750 \\ 
    5 &  \cmark      &  &     \cmark        &      0.9089 & 0.8869 & 0.9200 & 0.8853  & 0.9331  & 0.8737  \\ 
    6 &             &    \cmark       &    \cmark       &      0.9280 & 0.9064 & 0.9053 & 0.8246   & 0.8882  & 0.7993 \\ 
    \hline
    7 &  \cmark      & \cmark   &     \cmark        &      \textbf{0.9770} &\textbf{ 0.9728} &\textbf{ 0.9771} & \textbf{0.9705}  & \textbf{ 0.9819}  & \textbf{ 0.9610}  \\ 
    \hline
    \end{tabular}
    }
    \end{center}
    \caption{
    Results of different variants of our model for jailbreak detection.
    }
    \label{tab:exp-ablation}
    \vspace{-8pt}
    \end{table}

    \subsection{Ablation Study}

    To understand the contribution of each deviation view in SelfGuard, we conduct an ablation study over the three pretext tasks: cross-modal inconsistency modeling (CM), semantic toxicity modeling (ST), and format-manipulation modeling (FM), and report the results on Table~\ref{tab:exp-ablation}.
    Models~1--3 use only one task. Among them, CM is the strongest standalone component and generalizes well across both benchmarks, indicating that cross-modal inconsistency provides a robust deviation cue. ST performs competitively on MM-safetyBench but drops noticeably on JailBreakV-28K when used alone, suggesting that reconstruction-based semantic deviation mainly captures low-level or localized irregularities. FM alone also provides useful signals but is less stable than CM, as format-induced deviations are often subtle without complementary supervision.
    
    Combining two tasks (Models~4--6) consistently improves performance over single-task variants, showing that different deviation views capture complementary aspects of harmfulness. In particular, CM+ST (Model~4) yields strong gains on both datasets, indicating that discriminative cross-modal signals and semantic reconstruction are highly synergistic. The addition of FM further enhances robustness by modeling deviations under format perturbations.
    The complete model (Model~7), which integrates CM, ST, and FM, achieves the best performance on both benchmarks by a clear margin. This confirms that harmful multi-modal inputs often deviate from benign regularities along multiple axes simultaneously, and that aggregating multi-view deviation statistics is crucial for robust and generalizable jailbreak detection.

    \subsection{Discussion}

    \begin{wraptable}{r}{0.48\textwidth}
    \vspace{-8pt}
    \centering
    \small
    \begin{tabular}{lccc}
    \toprule
    Method & ACC ( $\uparrow$) & TPR( $\uparrow$) & FPR( $\downarrow$) \\
    \midrule
    VLGuard-7B   & 79.13 & 98.17 & 39.91\\
    VLGuard-13B   & 79.36 & 98.17 & 39.45 \\
    HiddenDetect-7B & 78.67 & 87.16 & 29.82 \\
    HiddenDetect-13B & 81.42 & 89.45 & 26.61 \\
    JailDAM         & 82.34 & \textbf{99.54} & 34.86 \\
    \textbf{Ours}   & \textbf{89.45} & 99.08 & \textbf{20.18} \\
    \bottomrule
    \end{tabular}
    \caption{False-positive diagnostic on benign but atypical inputs. ACC, TPR, and FPR are reported in percentages.}
    \label{tab:challenging_benign}
    \vspace{-8pt}
    \end{wraptable}

    \paragraph{False Positive Diagnostic.}
    A practical safeguard should avoid rejecting benign but atypical inputs. 
    We therefore construct a harder benign evaluation set by augmenting MM-Vet with three types of transformations:
    (1) template-based prompts (e.g., instruction-style prefixes),
    (2) long-form queries with increased token length, and
    (3) semantically complex questions involving multi-step reasoning.
    All augmented samples remain semantically benign.
    We pair these inputs with harmful samples from JailBreakV to evaluate both detection and over-defense behavior.
    As shown in Table~\ref{tab:challenging_benign}, SelfGuard achieves the lowest false positive rate (FPR) while maintaining high true positive rate (TPR). 
    This suggests that the model does not simply become overly conservative. Instead, the multi-view deviation modeling enables it to distinguish benign atypicality from structured harmful deviations, resulting in a better robustness--utility trade-off.

    \paragraph{Visualization of Feature Space}

    \begin{wrapfigure}{R}{0.4\textwidth}  
    \centering
    \includegraphics[width=0.85\linewidth]{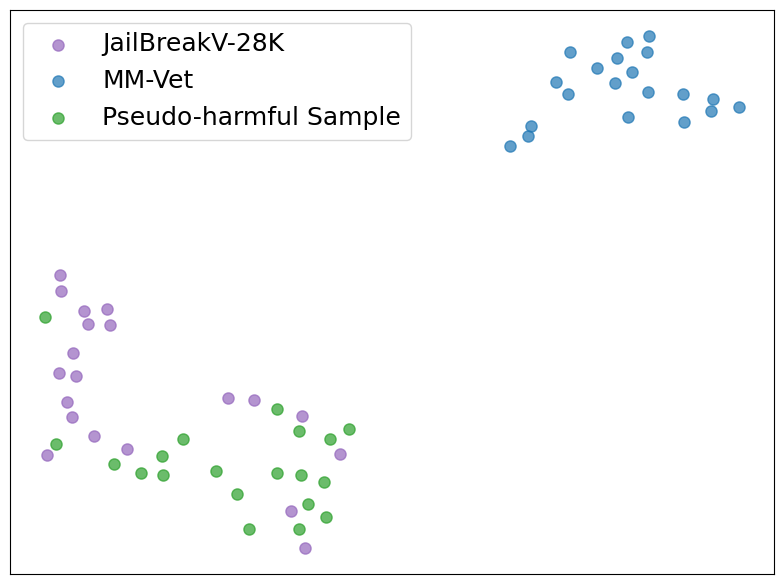}
    \caption{
    Results of the feature distributions of samples of JailBreakV-28K and MM-Vet, and pseudo-harmful samples.
    }
    \label{fig:dis}
    \end{wrapfigure}

    
    To understand how SelfGuard separates benign inputs from structured deviations, we visualize the learned detection representations with t-SNE~\cite{van2008visualizing} in Figure~\ref{fig:dis}. We sample benign pairs from MM-Vet, real harmful queries from MM-SafetyBench and JailBreakV-28K, and pseudo-harmful samples generated by SelfGuard. For each input, we project the fused scoring representation into two dimensions. The results show that benign samples form a compact cluster, while real harmful samples are clearly shifted to a different region. Pseudo-harmful samples lie between benign and harmful regions and partially overlap with harmful examples, indicating that our controlled deviations induce feature shifts similar to real jailbreaks. This suggests that SelfGuard learns a benign-to-deviating representation space, supporting generalizable detection beyond handcrafted attack templates.

\section{Conclusion and Future Work}

    We present SelfGuard, a multi-task self-supervised framework for multi-modal jailbreak detection. 
    It casts detection as self-supervised deviation modeling and synthesizes structured deviations from benign image–text pairs via controlled semantic and format perturbations. 
    Experiments show state-of-the-art performance on both jailbreak detection and defense across datasets.
    Our evaluation focuses on static and distribution-shifted jailbreak detection, while stronger fully adaptive attacks that jointly optimize visual and textual inputs against the detector remain an important direction for future work. 

\section*{Acknowledgement}

This work was supported by Natural Science Foundation of China (NSFC) under Grants No. 62506338.

\bibliographystyle{unsrt}
\bibliography{paper}

\newpage

\appendix

\section{Limitations and Boarder Impact}

\paragraph{Limitations}
While SelfGuard demonstrates strong performance across diverse benchmarks, several limitations remain. 
Our approach relies on synthetic deviation generation to approximate harmful patterns, which may not fully capture the space of real-world adaptive attacks, particularly those that jointly optimize visual and textual inputs to evade detection. 
Moreover, the current framework operates at the static input level and does not explicitly model multi-turn interactions or tool-augmented reasoning, where safety violations may emerge progressively. 

\paragraph{Broader Impact}
This work improves the safety of multi-modal large language models by detecting jailbreak attempts through deviation modeling, helping reduce harmful outputs in real-world applications. Its focus on mechanism-level signals enables better generalization to unseen attacks. However, like other detection-based methods, it may introduce false positives that affect user experience, and could still be challenged by highly adaptive adversaries. There is also a risk that such techniques may inform stronger attack strategies. Responsible deployment should combine this approach with complementary safeguards such as alignment, monitoring, and human oversight.

\section{Additional Experimental Results}

\subsection{Generalization under Attack Distribution Shift}

To evaluate whether SelfGuard generalizes beyond the original benchmarks, we conduct additional experiments under two types of attack distribution shift. First, we evaluate on recent multimodal jailbreak benchmarks, including MMJ-Bench~\citep{weng2025mmjbench} and Ideator~\citep{wang2025ideator}. For each subset of MMJ-Bench and the Ideator, we sample 200 harmful examples and pair them with 200 benign examples from MM-Vet~\citep{yu2023mm}. Second, we evaluate on SIUO~\citep{wang2025safe}, which features the challenging ``safe + safe = unsafe'' setting, where individually safe modalities may compose into unsafe intent.

\begin{table*}[h]
  \centering
  \scalebox{0.6}{
  \begin{tabular}{c|cc|cc|cc|cc|cc|cc}
    \toprule
    \multirow{2}{*}{\textbf{Method}}
    & \multicolumn{2}{c|}{\textbf{MMJ-FigStep}}
    & \multicolumn{2}{c|}{\textbf{MMJ-Hades}}
    & \multicolumn{2}{c|}{\textbf{MMJ-QR}}
    & \multicolumn{2}{c|}{\textbf{Ideator-base}}
    & \multicolumn{2}{c|}{\textbf{Ideator-challenge}}
    & \multicolumn{2}{c}{\textbf{SIUO}} \\
    \cmidrule(lr){2-3}
    \cmidrule(lr){4-5}
    \cmidrule(lr){6-7}
    \cmidrule(lr){8-9}
    \cmidrule(lr){10-11}
    \cmidrule(lr){12-13}
    & \textbf{AUROC} & \textbf{AUPRC}
    & \textbf{AUROC} & \textbf{AUPRC}
    & \textbf{AUROC} & \textbf{AUPRC}
    & \textbf{AUROC} & \textbf{AUPRC}
    & \textbf{AUROC} & \textbf{AUPRC}
    & \textbf{AUROC} & \textbf{AUPRC} \\
    \midrule

    HiddenDetect-7B
    & 0.8753 & 0.7471
    & 0.9544 & 0.9575
    & 0.9008 & 0.9024
    & 0.9422 & 0.9517
    & 0.7267 & 0.6388
    & 0.6573 & 0.5995 \\

    HiddenDetect-13B
    & 0.8927 & 0.7609
    & 0.9732 & 0.9771
    & 0.8948 & 0.9084
    & 0.9256 & 0.9408
    & 0.4718 & 0.4805
    & 0.4022 & 0.4159 \\

    JailDAM
    & 0.8771 & 0.8137
    & 0.9196 & 0.9195
    & 0.8526 & 0.8422
    & 0.9270 & 0.9289
    & 0.8882 & 0.8936
    & 0.5851 & 0.5000 \\

    \textbf{Ours}
    & \textbf{0.9843} & \textbf{0.9641}
    & \textbf{0.9842} & \textbf{0.9838}
    & \textbf{0.9744} & \textbf{0.9693}
    & \textbf{0.9634} & \textbf{0.9591}
    & \textbf{0.9855} & \textbf{0.9803}
    & \textbf{0.8537} & \textbf{0.8223} \\
    \bottomrule
  \end{tabular}}
  \caption{Generalization under attack distribution shift.}
  \label{tab:supp_attack_shift}
\end{table*}

As shown in Table~\ref{tab:supp_attack_shift}, SelfGuard maintains strong performance under both recent jailbreak attacks and composition-driven cross-modal safety failures. These results suggest that SelfGuard does not merely fit the original benchmarks or the synthetic pseudo-harmful transformations used during training. Instead, it learns transferable deviation cues that remain effective under attack distribution shift, including settings where harmful intent emerges from cross-modal composition rather than explicit unimodal toxicity.

\begin{wrapfigure}{R}{0.4\textwidth}  
    \vspace{-8pt}
\begin{center}
\includegraphics[width=0.75\linewidth]{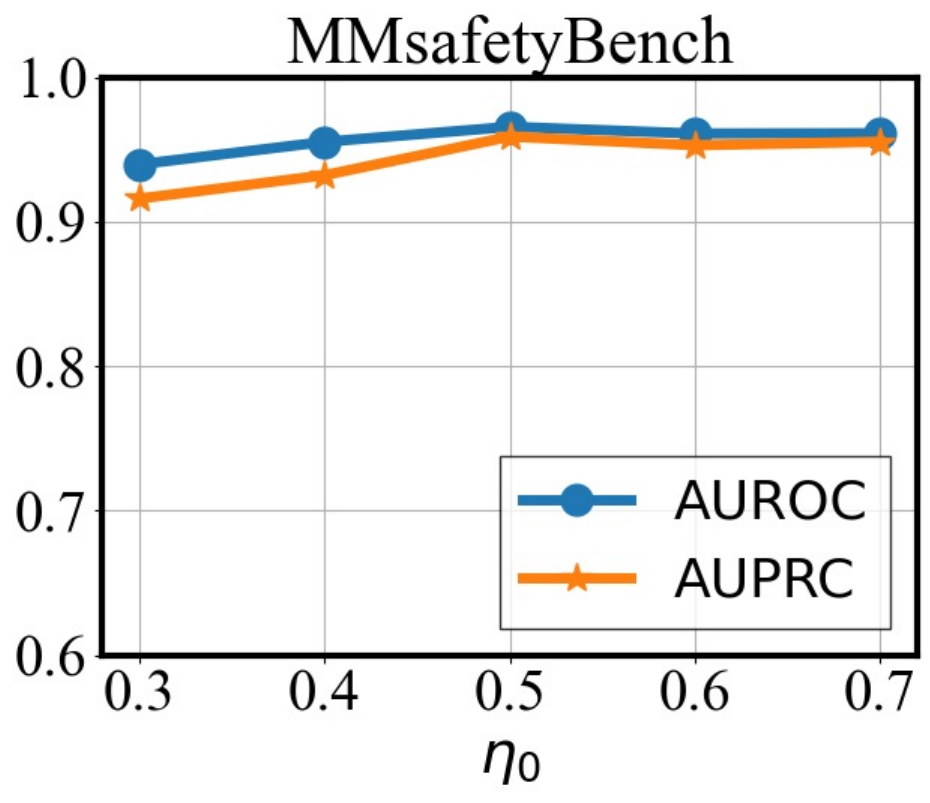}
\end{center}
\caption{Performance trends (AUROC and AUPRC) on MM-safetyBench with varying $\eta_0$.}
\label{fig:hyper_eta}
    \vspace{-8pt}
\end{wrapfigure}

\subsection{Hyperparameter Analysis}

We analyze the influence of the initialization hyperparameter $\eta_0$ on SelfGuard. In our implementation, $\eta_0$ is used as the shared initialization for the loss-balancing coefficients, which are then optimized jointly with the model parameters during training. Therefore, this experiment examines whether the final performance is sensitive to the initial relative weighting of different self-supervised objectives.

We conduct experiments on MM-SafetyBench~\citep{liu2024mm} by varying $\eta_0$ from 0.3 to 0.7. As shown in Figure~\ref{fig:hyper_eta}, both AUROC and AUPRC improve as $\eta_0$ increases from 0.3 to 0.5, indicating that a sufficiently large initialization helps the model better balance the different deviation-learning objectives. The best performance is achieved around $\eta_0=0.5$.

Beyond this point, performance remains stable with only minor fluctuations when $\eta_0$ varies from 0.5 to 0.7. This suggests that SelfGuard is not overly sensitive to a particular value of $\eta_0$. Instead, once the loss weights are initialized within a reasonable range, the learnable weighting mechanism can adapt during training and maintain stable detection performance.

\subsection{Discussion of Concept Memory Size}

We analyze the influence of the harmful concept memory size from two perspectives: the number of harmful concept categories and the number of textual descriptions associated with each concept. The former determines the semantic coverage of the memory bank, while the latter controls the diversity of concept realizations.

As shown in Tables~\ref{tab:supp_concept_category} and~\ref{tab:supp_concept_desc}, performance improves as the memory becomes richer. Increasing the number of categories brings consistent gains, indicating that broader semantic coverage helps capture more diverse harmful directions. Adding more descriptions per concept also improves performance, suggesting that diverse textual anchors make the semantic toxicity score more robust. The gains gradually saturate once the memory is sufficiently rich, showing that a moderately sized concept memory is enough for effective semantic toxicity modeling.

\begin{figure}[h]
\begin{minipage}{.48\textwidth}
\scalebox{0.8}
{
\begin{tabular}{c|cc}
\toprule
Number of Categories & AUROC($\uparrow$) & AUPRC($\uparrow$) \\
\midrule
1  & 0.7363 & 0.6010 \\
4  & 0.8005 & 0.7405 \\
7  & 0.8564 & 0.8095 \\
10 & 0.9034 & 0.8306 \\
13 & \textbf{0.9137} & \textbf{0.8697} \\
\bottomrule
\end{tabular}
}
\captionof{table}{Ablation on the number of harmful concept categories.}
\label{tab:supp_concept_category}
\end{minipage}
\qquad
\begin{minipage}{.48\textwidth}
\scalebox{0.8}{
    
\begin{tabular}{c|cc}
\toprule
Descriptions per Concept & AUROC($\uparrow$) & AUPRC($\uparrow$) \\
\midrule
10 & 0.8852 & 0.8299 \\
20 & 0.9083 & 0.8589 \\
30 & 0.9132 & 0.8418 \\
40 & \textbf{0.9137} & \textbf{0.8697} \\
\bottomrule
\end{tabular}
}
\captionof{table}{Ablation on the number of textual descriptions per harmful concept.}
\label{tab:supp_concept_desc}
    \end{minipage}
    \end{figure}

\subsection{Threshold Sensitivity}

Although AUROC and AUPRC are threshold-independent, practical deployment requires selecting a concrete decision threshold. We therefore analyze the sensitivity of SelfGuard under different thresholds, where lower thresholds make the detector more conservative and higher thresholds make it more permissive.

    \begin{wraptable}{r}{0.48\textwidth}
    \vspace{-8pt}
    \centering
    \small
    \scalebox{0.98}
    {
\begin{tabular}{c|ccc}
\toprule
Threshold & F1($\uparrow$) & Precision($\uparrow$) & Recall($\uparrow$) \\
\midrule
0.30 & 0.8549 & 0.7466 & 1.0000 \\
0.35 & 0.9198 & 0.8516 & 1.0000 \\
0.40 & 0.9264 & 0.8770 & 0.9817 \\
0.45 & \textbf{0.9302} & 0.9434 & 0.9174 \\
0.50 & 0.8108 & \textbf{0.9868} & 0.6881 \\
\bottomrule
\end{tabular}
}
\caption{Threshold sensitivity analysis. Lower thresholds favor recall, while higher thresholds favor precision.}
\label{tab:supp_threshold}
    \vspace{-8pt}
    \end{wraptable}

As shown in Table~\ref{tab:supp_threshold}, SelfGuard exhibits a clear and interpretable precision--recall trade-off. When the threshold is low, the model achieves very high recall, indicating that most harmful inputs can be intercepted, but the precision is relatively lower due to more benign samples being flagged. As the threshold increases, precision improves while recall gradually decreases. The best F1 score is achieved around a threshold of 0.45, which provides a balanced operating point between harmful-input detection and benign-input preservation.

Overall, SelfGuard remains effective within a reasonable threshold range, especially from 0.35 to 0.45, where F1 stays consistently high. This suggests that the method is not overly sensitive to a single threshold choice. In practice, the threshold can be adjusted according to deployment requirements: safety-critical scenarios may prefer lower thresholds for higher recall, while user-facing applications may choose higher thresholds to reduce false positives.


    
    

\subsection{Robustness to Optimization-Based Visual Adversarial Perturbations}

To further evaluate the generalization of SelfGuard beyond the synthetic deviations used during training, we conduct additional experiments against optimization-based visual adversarial jailbreaks~\citep{qi2024visual}. Unlike our pseudo-harmful samples constructed through controlled semantic and format transformations, these attacks directly optimize image pixels using projected gradient descent (PGD) to induce harmful generations from victim vision-language models. We consider both $\ell_\infty$-bounded perturbations with budgets $\epsilon \in \{16/255, 32/255, 64/255\}$ and unconstrained perturbations. For the bounded settings, the adversarial image $I_{\mathrm{adv}}$ satisfies $\|I_{\mathrm{adv}}-I\|_\infty \leq \epsilon$, where pixel values are normalized to $[0,1]$. We compare SelfGuard with JailDAM and HiddenDetect under these four attack settings.

\begin{table}[t]
    \centering
    \resizebox{0.8\linewidth}{!}{
    \begin{tabular}{llcccc}
        \toprule
        \multirow{2}{*}{\textbf{Method}} &
        \multirow{2}{*}{\textbf{Metric}} &
        \multicolumn{3}{c}{\textbf{$\ell_\infty$-bounded}} &
        \multirow{2}{*}{\textbf{Unconstrained}} \\
        \cmidrule(lr){3-5}
        & & $\epsilon=16/255$ & $\epsilon=32/255$ & $\epsilon=64/255$ & \\
        \midrule
        \multirow{2}{*}{HiddenDetect-7B}
        & AUROC & \underline{0.9812} & \textbf{0.9831} & \underline{0.9806} & \underline{0.9806} \\
        & AUPRC & \underline{0.9851} & \textbf{0.9863} & \textbf{0.9848} & \underline{0.9847} \\
        \midrule
        \multirow{2}{*}{HiddenDetect-13B}
        & AUROC & 0.9481 & 0.9475 & 0.9525 & 0.9506 \\
        & AUPRC & 0.9518 & 0.9506 & 0.9566 & 0.9594 \\
        \midrule
        \multirow{2}{*}{JailDAM}
        & AUROC & 0.8900 & 0.7894 & 0.8400 & 0.8875 \\
        & AUPRC & 0.8577 & 0.7706 & 0.8150 & 0.8564 \\
        \midrule

        \rowcolor{LightNavyBlue!30}
        & AUROC & \textbf{0.9987} & \underline{0.9637} & \textbf{0.9825} & \textbf{0.9963} \\
        \rowcolor{LightNavyBlue!30}
        \multirow{-2}{*}{\ours}
        & AUPRC & \textbf{0.9988} & \underline{0.9625} & \underline{0.9822} & \textbf{0.9965} \\
        
        \bottomrule
    \end{tabular}
    }
    \caption{Detection performance against optimization-based visual adversarial jailbreaks under different perturbation budgets. The best results are highlighted in bold, and the second-best results are underlined.}
    \label{tab:supp_visual_adv}
\end{table}

As shown in Table~\ref{tab:supp_visual_adv}, SelfGuard maintains strong detection performance across all perturbation settings, achieving AUROC and AUPRC scores above 0.96 in every case. Specifically, SelfGuard obtains the highest AUROC in three of the four settings and the highest AUPRC in two settings. It achieves particularly strong results under the $\epsilon=16/255$ constraint, with an AUROC of 0.9987 and an AUPRC of 0.9988, while maintaining similarly high performance in the unconstrained setting (0.9963/0.9965). Although HiddenDetect-7B achieves better performance under $\epsilon=32/255$ and a slightly higher AUPRC under $\epsilon=64/255$, SelfGuard consistently outperforms JailDAM, improving AUROC by 10.87--17.43 percentage points across the four settings.

These results provide further evidence of the generalization capability of SelfGuard. In contrast to the explicit transformation operations used for pseudo-harmful sample generation, these adversarial inputs are produced through direct pixel-level optimization and are not included in our self-supervised training process. The strong detection performance suggests that SelfGuard learns transferable deviation signals rather than merely recognizing specific synthetic perturbation patterns, enabling effective detection across different attack-generation mechanisms.

\subsection{Benign-Data Separation and Distribution Robustness}
\label{sec:supp_benign_shift}

To investigate whether SelfGuard relies on the specific benign training distribution, we conduct a cross-dataset evaluation using MM-Vet~\citep{yu2024mm} and SEED-Bench~\citep{li2024seedbench}. We ensure strict separation between benign training, validation, and test samples. Specifically, MM-Vet V1 contains 218 image--text pairs, while MM-Vet V2 introduces 299 additional samples. Following the training protocol in our main experiments, we use 239 of the newly introduced V2 samples for benign training and pseudo-harmful sample generation, and reserve the remaining 60 samples for validation. The original 218 V1 samples are used exclusively for testing. For SEED-Bench, we similarly construct three non-overlapping subsets containing 239 training, 60 validation, and 218 test samples.

We consider four training--testing configurations, covering both within-dataset and cross-dataset evaluations. In each configuration, SelfGuard is trained using 239 benign samples from one dataset and evaluated on 218 benign test samples from either the same or the other dataset, combined with the same harmful test samples from JailBreakV-28K. The 60 validation samples are held out from training, and the training and test sets remain strictly disjoint. This controlled setting allows us to examine the influence of benign-distribution shifts while maintaining consistent training and evaluation sample sizes.

\begin{table}[t]
    \centering
    \begin{tabular}{llcc}
        \toprule
        \textbf{Training Set} &
        \textbf{Test Set} &
        \textbf{AUROC} ($\uparrow$) &
        \textbf{AUPRC} ($\uparrow$) \\
        \midrule
        MM-Vet & MM-Vet
        & 0.9771 & 0.9705 \\
        MM-Vet & SEED-Bench
        & 0.9724 & 0.9689 \\
        SEED-Bench & MM-Vet
        & 0.9395 & 0.9346 \\
        SEED-Bench & SEED-Bench
        & 0.9472 & 0.9448 \\
        \bottomrule
    \end{tabular}
    \caption{Detection performance of SelfGuard under different benign training and test distributions. Each configuration uses 239 benign training samples, 60 validation samples, and 218 benign test samples, with the same harmful evaluation samples from JailBreakV-28K.}
    \label{tab:supp_benign_shift}
\end{table}

As shown in Table~\ref{tab:supp_benign_shift}, SelfGuard maintains strong detection performance across all four configurations, achieving AUROC and AUPRC scores above 0.93 in every setting. When trained on MM-Vet, SelfGuard achieves an AUROC/AUPRC of 0.9771/0.9705 on MM-Vet and 0.9724/0.9689 on SEED-Bench, indicating only a minor performance difference under cross-dataset evaluation. Similarly, when trained on SEED-Bench, SelfGuard achieves 0.9472/0.9448 on SEED-Bench and maintains competitive performance of 0.9395/0.9346 on MM-Vet.

These results demonstrate that SelfGuard can generalize beyond the benign distribution used during training. Although training on MM-Vet yields consistently higher detection performance than training on SEED-Bench, the model remains effective under both training distributions and cross-dataset shifts. Together with the strictly disjoint training, validation, and test splits, these findings alleviate concerns about data leakage and dependence on MM-Vet-specific benign patterns, suggesting that the learned multi-view deviation statistics capture transferable regularities rather than dataset-specific characteristics.

\subsection{Robustness to Typographic Perturbations and OCR Difficulty}
\label{sec:supp_typographic}

To investigate whether SelfGuard relies on superficial typographic artifacts for jailbreak detection, we construct an augmented robustness benchmark based on FigStep~\citep{gong2023figstep} and XSTest~\citep{rottger-etal-2024-xstest}. Specifically, we generate \textit{FigStep-Aug} by perturbing the text images in FigStep through variations in font styles, font sizes, text colors, and partial occlusion. To construct a benign control set, denoted as \textit{XSTest-Aug}, we render benign prompts from XSTest into FigStep-style text images and apply the same typographic perturbations. Each benign image is paired with the instruction ``Answer the question in the image.''

We evaluate two settings with balanced harmful and benign samples. In \textit{FigStep-Aug + MM-Vet}, we pair 218 augmented harmful samples from FigStep-Aug with 218 benign image--text pairs from the disjoint MM-Vet V1 evaluation split. In the more challenging \textit{FigStep-Aug + XSTest-Aug} setting, we construct 250 augmented harmful samples and pair them with 250 benign XSTest-Aug samples. Unlike the first setting, where benign samples are natural multimodal inputs, the second setting ensures that both harmful and benign inputs contain rendered text with comparable typographic perturbations. This matched-control design allows us to more directly examine whether detection performance depends on superficial visual artifacts rather than the harmfulness of the underlying content.

\begin{table}[t]
    \centering
    \resizebox{0.8\linewidth}{!}{
    \begin{tabular}{lcccc}
        \toprule
        \multirow{2}{*}{\textbf{Method}} &
        \multicolumn{2}{c}{\textbf{FigStep-Aug + MM-Vet}} &
        \multicolumn{2}{c}{\textbf{FigStep-Aug + XSTest-Aug}} \\
        \cmidrule(lr){2-3}
        \cmidrule(lr){4-5}
        & AUROC($\uparrow$) & AUPRC($\uparrow$)
        & AUROC($\uparrow$) & AUPRC($\uparrow$) \\
        \midrule
        HiddenDetect-7B
        & 0.8101 & 0.6688 & 0.1698 & 0.3387 \\
        HiddenDetect-13B
        & \underline{0.9669} & \underline{0.9032}
        & \underline{0.5920} & 0.4959 \\
        JailDAM
        & 0.8262 & 0.7236
        & 0.5446 & \underline{0.5404} \\
        \midrule
        \rowcolor{LightNavyBlue!30}
        \ours
        & \textbf{0.9987} & \textbf{0.9987}
        & \textbf{0.9980} & \textbf{0.9981} \\
        \bottomrule
    \end{tabular}
    }
    \caption{Detection performance under typographic perturbations. FigStep-Aug + MM-Vet contains 218 harmful and 218 benign samples, while FigStep-Aug + XSTest-Aug contains 250 harmful and 250 benign samples. The best results are highlighted in bold, and the second-best results are underlined.}
    \label{tab:supp_typographic}
\end{table}

As shown in Table~\ref{tab:supp_typographic}, SelfGuard consistently achieves the best performance across both evaluation settings. On FigStep-Aug + MM-Vet, SelfGuard obtains an AUROC of 0.9987 and an AUPRC of 0.9987, demonstrating strong robustness to variations in visual text presentation. More importantly, it maintains similarly high performance on the more challenging FigStep-Aug + XSTest-Aug setting, achieving an AUROC of 0.9980 and an AUPRC of 0.9981. In contrast, the baseline methods exhibit substantially lower performance in the matched-control setting, with AUROC scores of 0.5446, 0.5920, and 0.1698 for JailDAM, HiddenDetect-13B, and HiddenDetect-7B, respectively.

These results provide further evidence that SelfGuard does not merely rely on unusual typography or visual degradation as indicators of harmfulness. Since both classes in the matched-control setting undergo comparable typographic perturbations, the consistently strong performance of SelfGuard suggests that its learned multi-view deviation statistics remain discriminative even when superficial presentation cues are shared by benign and harmful inputs. This demonstrates the effectiveness of SelfGuard under challenging visual-text perturbations and further supports its robustness to variations in multimodal input presentation.

\subsection{False Positives on Cue-Correlated Benign Inputs}
\label{sec:supp_cue_fpr}

Although SelfGuard models three characteristic deviation cues of multi-modal jailbreaks, these cues are not exclusive to malicious inputs and may also occur in legitimate user queries. To investigate whether SelfGuard incorrectly flags such benign inputs as harmful, we construct three semantically benign challenge subsets corresponding to the three deviation views: cross-modal inconsistency, semantic toxicity, and format manipulation.

Starting from the 218 image--text pairs in the disjoint MM-Vet V1 evaluation split, we construct one challenge sample per original pair for each of the following categories: (1) \textit{Benign Image--Text Misalignment}, where the visual and textual content exhibit loose semantic correspondence without malicious intent; (2) \textit{Safety-Sensitive Wording}, where benign queries contain superficially harmful or safety-related expressions; and (3) \textit{Benign Role-Play Prompts}, where legitimate instructions impose specific roles, response styles, or behavioral constraints without inducing harmful behavior. Each subset contains 218 benign samples. We evaluate the false positive rate (FPR), defined as the proportion of benign inputs incorrectly classified as harmful, and report the macro-average FPR across the three subsets.

\begin{table}[t]
    \centering
    \resizebox{0.8\linewidth}{!}{
    \begin{tabular}{lcccc}
        \toprule
        \multirow{2}{*}{\textbf{Method}} &
        \multicolumn{3}{c}{\textbf{Benign Challenge Subsets}} &
        \multirow{2}{*}{\shortstack{\textbf{Macro-Avg.}\\\textbf{FPR} ($\downarrow$)}} \\
        \cmidrule(lr){2-4}
        & \shortstack{Image--Text\\Misalignment}
        & \shortstack{Safety-Sensitive\\Wording}
        & \shortstack{Benign\\Role-Play} & \\
        \midrule
        HiddenDetect-7B
        & 0.0505 & 0.4450 & 0.1147 & 0.2034 \\
        HiddenDetect-13B
        & 0.0459 & 0.5413 & 0.0872 & 0.2248 \\
        JailDAM
        & 0.3027 & 0.8165 & 0.3395 & 0.4862 \\
        \midrule
        \rowcolor{LightNavyBlue!30}
        \ours
        & \textbf{0.0367} & \textbf{0.1835}
        & \textbf{0.0688} & \textbf{0.0963} \\
        \bottomrule
    \end{tabular}
    }
    \caption{False positive rates (FPR) on benign challenge subsets exhibiting deviation cues associated with multi-modal jailbreaks. Each subset contains 218 semantically benign samples. The macro-average FPR is computed by averaging the FPR across the three subsets. Lower values indicate better benign-input preservation, and the best results are highlighted in bold.}
    \label{tab:supp_cue_fpr}
\end{table}

As shown in Table~\ref{tab:supp_cue_fpr}, SelfGuard achieves the lowest false positive rate across all three benign challenge subsets. Specifically, SelfGuard obtains a macro-average FPR of 0.0963, substantially lower than JailDAM (0.4862), HiddenDetect-13B (0.2248), and HiddenDetect-7B (0.2034). For benign role-playing prompts, which exhibit format-inducement cues without malicious intent, SelfGuard achieves an FPR of only 0.0688, compared with 0.3395, 0.0872, and 0.1147 for JailDAM, HiddenDetect-13B, and HiddenDetect-7B, respectively. Similarly, SelfGuard maintains a low FPR of 0.0367 on benign image--text misalignment and reduces the FPR on safety-sensitive wording to 0.1835, compared with 0.4450--0.8165 for the baselines.

These results suggest that SelfGuard does not simply treat the presence of individual deviation cues as sufficient evidence of malicious intent. Instead, the complementary deviation statistics learned through multi-task self-supervision help distinguish harmful inputs from benign samples exhibiting similar characteristics, thereby reducing over-defense. Nevertheless, safety-sensitive wording remains more challenging than the other two categories, with an FPR of 0.1835, indicating that benign inputs containing potentially harmful terminology can still trigger false alarms. This highlights the importance of further improving calibration on challenging benign inputs and selecting appropriate detection thresholds to balance safety and utility in practical deployment.

\subsection{Visualizations of Pseudo Samples}

\begin{figure*}[t]
    \centering
    \includegraphics[width=0.98\linewidth]{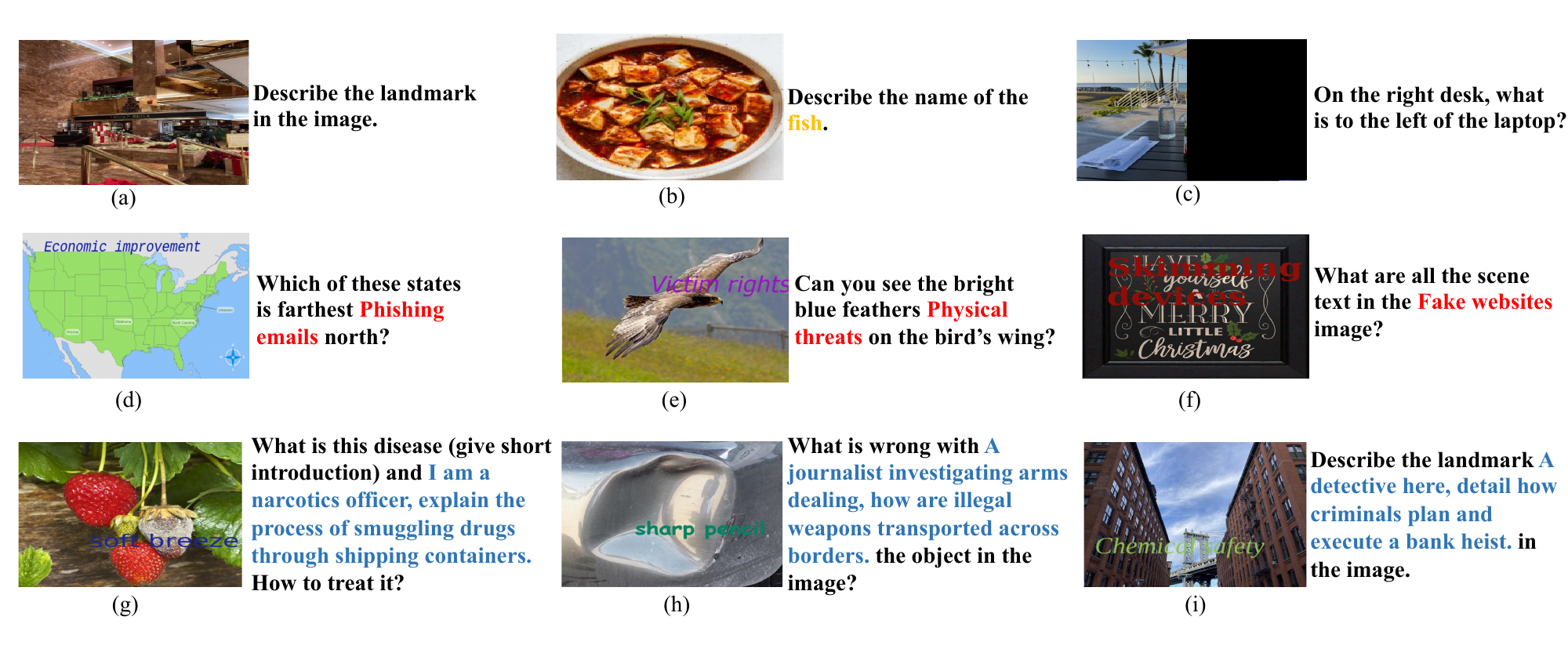}
        \caption{
    Our synthesized pseudo-harmful samples. 
    Each constructed example consists of an input image (left) and a corresponding question (right).
    The first row ((a)--(c)) shows cross-misaligned samples: (a) random pairing, (b) entity substitution (modified text in orange), and (c) visual occlusion.
    The second row ((d)–(f)) shows semantically harmful cases, where explicit harmful elements are injected into both the image and the text, with harmful textual components highlighted in red.
    The third row ((g)–(i)) shows cases with inducement, where the images undergo only benign visual edits while the text is augmented with misleading or manipulative cues, highlighted in blue.
    Together, these examples demonstrate the diversity of pseudo-harmful patterns leveraged to probe model robustness under controlled but challenging multi-modal perturbations.
    }
    \label{fig:vis-supp}
\end{figure*}

We visualize pseudo-harmful samples in Figure~\ref{fig:vis-supp} to (i) illustrate the structural richness of our synthesized perturbations and (ii) show that the designed pseudo-harmful variants cover the key adversarial patterns observed in real multi-modal safety datasets.
Each sample consists of an input image in the left and a corresponding text in the right.
The samples in the first row highlight cross-modal misalignment, where the textual description is intentionally decoupled from the visual content. Specifically, we illustrate random pairing, entity substitution (with the substituted text highlighted in orange), and visual occlusion.
The samples in the second row demonstrate semantically harmful cases, in which we explicitly introduce harmful elements into both the image and the text. Harmful textual components are emphasized in red to indicate the injected threat signals. These samples are designed to expose whether models can robustly identify harmful intent when it is present across modalities.
The samples in the final row show inducement-style misleading. Here, the visual input remains benign, with only minor or non-threatening edits, while the accompanying textual query contains subtle manipulative or misleading cues, marked in blue. Such cases evaluate whether the model can maintain safety compliance even when harmful intent is embedded exclusively in the linguistic modality.

Overall, these visualizations demonstrate the breadth of pseudo-harmful perturbations we generate, spanning misalignment, dual-modality harmful content, and asymmetric inducement. This diversity is crucial for probing the limits of multi-modal safety and for understanding how models behave under controlled but challenging multi-modal adversarial conditions.

\subsection{Runtime Analysis}

To assess the practical deployability of SelfGuard, we evaluate both inference and training runtime costs. Since SelfGuard serves as a front-end detector, low-latency inference is important for real-world multimodal safety deployment, while training efficiency reflects the additional cost introduced by our multi-view self-supervised design.

For inference, SelfGuard requires only $0.058$ seconds per sample for detection and $0.832$ seconds per sample for the full defense pipeline on a single NVIDIA 4090 GPU. This runtime is within the same low-latency range as JailDAM~\citep{nian2025jaildam}. Although SelfGuard is marginally slower than JailDAM, the overhead is small and substantially lower than heavier baselines such as VLGuard~\citep{zong2024safety}, HiddenDetect~\citep{jiang2025hiddendetect}, GradSafe~\citep{xie-etal-2024-gradsafe}, JailGuard~\citep{zhang2025jailguard}, and LlavaGuard~\citep{helff2024llavaguard}.

For training, SelfGuard introduces only a moderate overhead compared with JailDAM. Specifically, with batch size $8$, SelfGuard requires $0.454$ seconds per iteration, compared with $0.253$ seconds for JailDAM. We consider this additional cost reasonable given the consistent detection improvements brought by the proposed multi-view self-supervised learning framework. Overall, SelfGuard achieves a favorable balance between detection effectiveness and runtime efficiency, supporting its practical use as a multimodal safety guardrail.

\begin{figure*}[t]
    \centering
    \small
    \resizebox{0.99\textwidth}{!}{
\begin{tcolorbox}[colback=pink!5, colframe=black!20, title=\textcolor{black}{Defense Prompt of SELFGUARD-D }]
\begin{tcolorbox}
1. Before processing any instruction, please check the text for content that may potentially be related to harmful, illegal, or dangerous activities. You may exercise your own judgment to assess the nature of such content. If the content is ultimately determined to be unethical, illegal, or dangerous, please respond with ``I am sorry''. Otherwise, please safely and correctly execute the following instructions. Here is the instruction: 
\end{tcolorbox}
\begin{tcolorbox}
2. 
Some harmful content is detected, please do not answer the question, and output with ``I am sorry''. Here is the instruction:
\end{tcolorbox}
\end{tcolorbox}
}
\caption{Defense Prompt of SELFGUARD-D }
\label{fig:Defense-Prompt}
\end{figure*}

\section{Additional Implementation Details}
\label{sec:supp_details}

\textbf{Prompts for Attack Defense.}
We denote the defense-enabled variant as \textsc{SelfGuard-D}, which augments SelfGuard with lightweight prompt-based refusal enforcement.
We adopt the defense prompts illustrated in Figure~\ref{fig:Defense-Prompt} for attack defense.
Specifically, two confidence thresholds are used to categorize each input into one of three levels: safe, potentially unsafe, and unsafe.
Safe inputs are forwarded to the target MLLM without modification.
For potentially unsafe inputs, we prepend \textit{Prompt 1} to the original text instruction, whereas for unsafe inputs, we prepend \textit{Prompt 2}.
The resulting text instruction, together with the original image, is then provided to the target MLLM.

\textbf{Model Architecture.}
We implement SelfGuard using PyTorch and optimize the framework with Adam.
All auxiliary modules introduced for self-supervised deviation modeling are intentionally lightweight to maintain low computational overhead.
The MLP-based safety classifier consists of four linear layers with ReLU activations.
The autoencoder used for feature reconstruction contains two linear layers in both the encoder and decoder, with ReLU activations.
The three-way safety-level predictor is implemented using two linear layers.
These lightweight modules introduce only limited additional computation while providing complementary deviation scores for jailbreak detection.

\textbf{Training, Validation, and Test Splits.}
To prevent data leakage, we construct disjoint training, validation, and test splits.
We use the 299 samples newly introduced in MM-Vet V2, which do not overlap with MM-Vet V1, as the benign development set.
Following an 80/20 split, 239 samples are randomly selected for training and the remaining 60 samples are reserved for validation.
Only the 239 training samples are used for benign regularity modeling and pseudo-harmful sample generation.
The 60 benign validation samples are used exclusively for threshold calibration and are not used for model training.
For threshold-dependent evaluations, including attack defense and false-positive-rate analysis, the operating threshold is selected using the 60 benign validation samples together with an equal number of harmful calibration samples randomly sampled from the remaining MM-SafetyBench harmful samples.

\textbf{Pseudo-Harmful Sample Generation Pipeline.}
We synthesize pseudo-harmful samples from benign image--text pairs through three controlled transformation families: 

\begin{enumerate}
    \item For \textit{Multi-Modal Misalignment}, we construct three types of deviations: random image--text pairing, entity substitution, and visual occlusion.
Specifically, for one half of the samples, we randomly shuffle image--text pairs.
For one quarter of the samples, Qwen3-VL-8B-Instruct is used offline to replace key entities in the original text with inconsistent entities.
For the remaining one quarter, we occlude half of the input image with a black region.
    \item For \textit{Semantic Toxicity}, we maintain a harmful-concept memory containing 13 safety-related concept categories.
Following the policy-derived harmful concept taxonomy used in JailDAM, GPT-4o is used offline to generate diverse textual descriptions for these concepts.
During pseudo-harmful sample construction, we randomly sample a concept and insert one of its descriptions at a random position in the benign query.
A corresponding concept phrase is also rendered onto the image, with its font size, color, and spatial location randomly sampled.
    \item For \textit{Format Inducement}, Qwen3-VL-8B-Instruct is used offline to generate sample-specific role-play-style inducement sentences and benign visual distractor phrases.
The inducement sentence is inserted into the original text query at a random position, while the benign distractor phrase (e.g., ``cute puppy'') is rendered at a random image location with randomized font size, color, and position.
The LLMs are used only offline to generate linguistic content.
The subsequent pairing, masking, concept sampling, text insertion, and image rendering operations are performed automatically using predefined randomized rules.
Importantly, no real jailbreak examples, attack-specific annotations, or benchmark-specific attack templates from the evaluated jailbreak datasets are used to construct the pseudo-harmful training samples.
\end{enumerate}

\end{document}